%% file: main.tex
\documentclass[letterpaper, 10 pt, conference]{ieeeconf}  % Comment this line out if you need a4paper

\IEEEoverridecommandlockouts                              % This command is only needed if 
\usepackage{tikz}
\usetikzlibrary{shapes.geometric, positioning, calc}
\usepackage{graphicx}
\usepackage{amsmath} 
\usepackage{amssymb}
\usepackage{cite}
\usepackage{tikz}
\usepackage{amsmath}
\usepackage{tabularx}
\usepackage{array}
\usepackage{booktabs}
\usepackage{bm}
\usepackage{algorithm}
\usepackage{algorithmic}
\usepackage{bbm}
\usepackage{subcaption}
\PassOptionsToPackage{hyphens}{url}\usepackage{hyperref}
\usepackage{lipsum}
\usepackage{nicematrix}
\usepackage{fontawesome5}
\usepackage{pifont}% http://ctan.org/pkg/pifont
\usepackage{tablefootnote}

\DeclareMathOperator*{\mode}{mode}
\newcommand{\xmark}{\ding{55}}%

\graphicspath{{figures/}} % import figures directory

\title{\bf
VANP: Learning Where to See for Navigation with \\ Self-Supervised Vision-Action Pre-Training
}

\author{Mohammad Nazeri, Junzhe Wang, Amirreza Payandeh, and Xuesu Xiao \\
\thanks{All authors are with the Department of Computer Science, George Mason University {\tt\scriptsize \{mnazerir, jwang69, apayande, xiao\}@gmu.edu}}
}

\begin{document}
\maketitle
\thispagestyle{empty}
\pagestyle{empty}

\begin{abstract}
\input{sections/abstract}
\end{abstract}

\section{Introduction}\label{sec:intro}
\input{sections/introduction}

\section{Related Work}\label{sec:related}
\input{sections/related_work}

\section{Methodology}\label{sec:method}
\input{sections/methodology}

\section{Experimental Results}\label{sec:results}
\input{sections/results}

\section{Conclusions and Future Work}\label{sec:conclusion}
\input{sections/conclusion}

\section{Acknowledgments}
This work has taken place in the RobotiXX Laboratory at George Mason University. RobotiXX research is supported by National Science Foundation (NSF, 2350352), Army Research Office (ARO, W911NF2220242, W911NF2320004, W911NF2420027), US Air Forces Central (AFCENT), Google DeepMind (GDM), Clearpath Robotics, and Raytheon Technologies (RTX).

\bibliographystyle{IEEEtran}
\bibliography{IEEEabrv,bibliography}
\end{document}

%% file: sections/abstract.tex
Humans excel at efficiently navigating through crowds without collision by focusing on specific visual regions relevant to navigation. However, most robotic visual navigation methods rely on deep learning models pre-trained on vision tasks, which prioritize salient objects---not necessarily relevant to navigation and potentially misleading. Alternative approaches train specialized navigation models from scratch, requiring significant computation. On the other hand, self-supervised learning has revolutionized computer vision and natural language processing, but its application to robotic navigation remains underexplored due to the difficulty of defining effective self-supervision signals. Motivated by these observations, in this work, we propose a Self-Supervised \underline{V}ision-\underline{A}ction Model for Visual \underline{N}avigation \underline{P}re-Training (VANP). Instead of detecting salient objects that are beneficial for tasks such as classification or detection, VANP learns to focus only on specific visual regions that are relevant to the navigation task. To achieve this, VANP uses a history of visual observations, future actions, and a goal image for self-supervision, and embeds them using two small Transformer Encoders. Then, VANP maximizes the information between the embeddings by using a mutual information maximization objective function. We demonstrate that most VANP-extracted features match with human navigation intuition. VANP achieves comparable performance as models learned end-to-end with half the training time and models trained on a large-scale, fully supervised dataset, i.e., ImageNet, with only 0.08\% data. 

%% file: sections/introduction.tex
In recent years, imitation learning, particularly behavior cloning~\cite{Pomerleau1988}, has become a leading approach for visual navigation models~\cite{Bojarski2016, Codevilla2018, Codevilla2019, Ohn-Bar2020, Chen2020c, Nazeri2021, Xiao2022b}. However, the performance of these models heavily relies on the visual features extracted by the model's visual encoder. Although the limited memory and processing power onboard robots restrict the size of models deployable in real time, with such limitations we still need accurate and efficient onboard visual encoders, making convolutional neural networks (CNNs) more desirable than larger Vision Transformer models (ViTs)~\cite{Dosovitskiy2021}.  

\begin{figure}[!t]
	\centering
        \includegraphics[width=\linewidth]{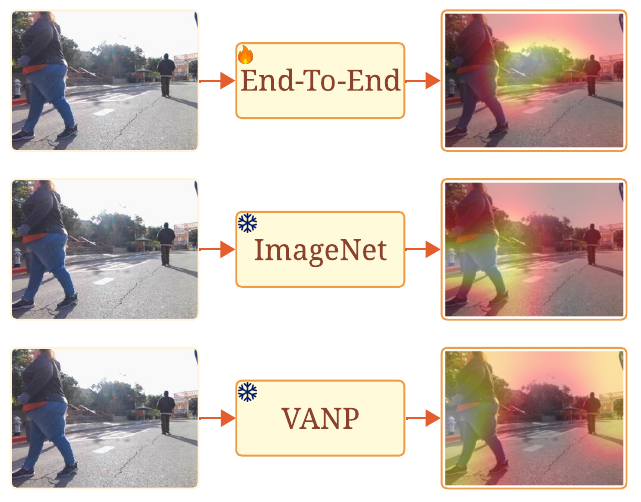}
	\caption{\textbf{Comparison of Activation Maps Learned by End-to-End, ImageNet, and VANP.} VANP can extract multiple regions of interest for navigation without downstream navigation supervision compared to single salient regions by End-to-End and ImageNet pre-trained models.}
	\label{fig:intro}
\end{figure}

Training a visual navigation-specific encoder from scratch requires a large amount of data, leading to high computational demands and extended training times~\cite{shah2022gnm, Shah2023vint}. To reduce this computational burden, most approaches use pre-trained vision models~\cite{Codevilla2019, Ohn-Bar2020}. While these models provide a decent scene representation, they specialize in extracting salient features for vision tasks such as object classification and detection~\cite{Deng2009}. These features may not always align with what is crucial for navigation~\cite{Vishniakov2024}. For example, following sidewalks, avoiding grass, or navigating around stairs and guardrails are essential for robots, but these features might not be captured by encoders trained for generic vision tasks. Consequently, pre-trained models, like those trained on ImageNet, can sometimes lead to navigation failures by focusing on irrelevant distractions~\cite{shah2022gnm, Shah2023vint}.

Self-Supervised Learning (SSL)~\cite{Chen2020b, Zbontar2021, Oquab2023, Bardes2023} has shown success in various computer vision tasks by extracting general features adaptable to downstream tasks with/without fine-tuning. For instance, Deep Neural Networks (DNNs) can be trained to predict the rotation of an image~\cite{Gidaris2018} or to reconstruct an image from its corrupted/obstructed version~\cite{He2022}. By completing these pretext tasks, DNNs learn to extract meaningful features from the data, which can be used to solve downstream tasks such as image classification and object detection~\cite{Balestriero2023}. 
However, a discrepancy exists between features extracted from generic models and those specifically needed for navigation. This leads us to ask the question: \emph{can we train visual encoders that extract only navigation-relevant features using self-supervision?}

Considering both the success of SSL on a variety of computer vision tasks and the oftentimes mismatched features provided by generic SSL models for navigation tasks, we present \underline{V}ision-\underline{A}ction \underline{N}avigation \underline{P}retraining (VANP), a non-contrastive self-supervised approach that completely relies on a navigation-specific pretext task to train the visual encoder without the need for negative samples.

The core idea behind VANP is inspired by how humans navigate in crowded spaces. We do not need to pay attention to all the people and objects in the scene, but only the ones that affect our navigation trajectory. To this end, VANP embeds visual history, future actions, and visual goal as self-supervision signals and leverages Transformers with additional context tokens (inspired by Bert~\cite{Devlin2019} and VisionTransformers~\cite{Dosovitskiy2021}) to generate embeddings. Then, VANP utilizes VICReg~\cite{Bardes2022} as the pretext objective function to maximize the mutual information between the embeddings. The trained visual encoder can therefore discard redundant features unnecessary for navigation and focus only on navigation-relevant regions. For example, Fig.~\ref{fig:intro} shows the activation map of the last layer of ResNet-50~\cite{He2015} trained with different methods. VANP learns navigation-relevant visual features with the help of our navigation-specific self-supervision signals. 

Our experimental results suggest that VANP-extracted features trained on a dataset~\cite{Karnan2022a} that only contains 0.08\% samples compared to ImageNet are as informative for a downstream navigation task as using ImageNet features.
The contributions of this work can be summarized as follows:
 \begin{itemize}
    \item An SSL framework to train a visual encoder for robotic navigation tasks;
    \item Insights into what is happening inside CNNs during navigation using different approaches; and
    \item A benchmark on short and long-term navigation interaction to show the performance of different approaches. 
\end{itemize}

%% file: sections/related_work.tex
Recent advances in natural language processing and computer vision, particularly those driven by self-supervised learning (SSL), motivate our work. In this section, we first compare common SSL approaches and then categorize applications of SSL into two groups for robotics and review their related works.

\textbf{Self-Supervised Learning:} SSL has shown promising results in recent years by almost reaching the performance of supervised baselines~\cite{Zbontar2021, Bardes2022}. Within SSL, two primary approaches have emerged: contrastive methods and information maximization methods. Both methodologies benefit from the use of the Siamese network architecture~\cite{Bromley1993}. Contrastive methods~\cite{He2020, Chen2020b} typically require large data batches and leverage loss functions designed to explicitly push dissimilar data points away from the representations of similar data. Consequently, the performance of these methods is highly dependent on the quality and quantity of negative samples~\cite{sikand2022visual}. Recent advancements have led to the development of contrastive approaches that do not necessitate negative samples for learning effective embeddings. These methods employ various strategies to achieve comparable performance, such as BYOL~\cite{Grill2020}, which utilizes a momentum encoder where one head receives a low-pass filtered version of the other. Alternatively, SimSiam~\cite{Chen2020a} achieves similar results by halting gradient flow within one of the heads.

\begin{figure*}[!t]
	\centering
        \includegraphics[width=\linewidth]{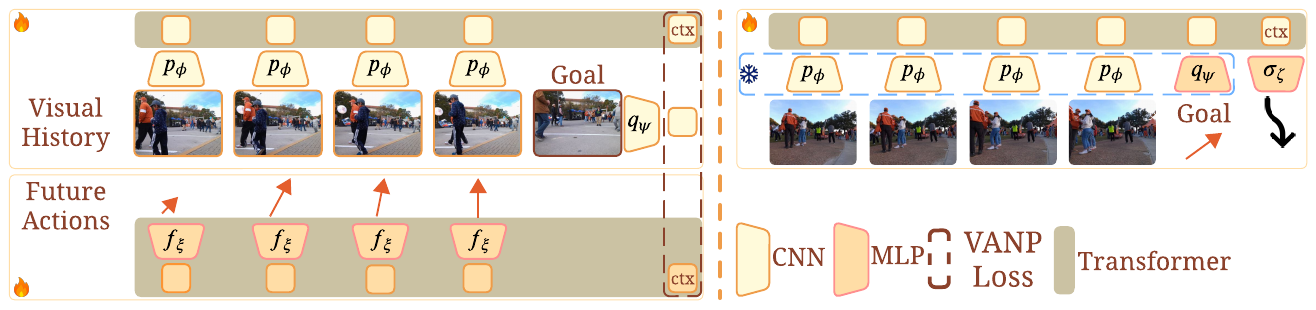}
	\caption{\textbf{VANP Architecture.} VANP learns to embed temporal features into spatial features by using a sequence of images and leveraging two TransformerEncoders with context tokens. VANP's loss maximizes the mutual information between history, future actions, and the goal (left). Then, by appending an MLP to the Transformer context token, VANP predicts future trajectories during the downstream navigation task (right).}
	\label{fig:arch}
\end{figure*} 

Information maximization methods such as BarlowTwins~\cite{Zbontar2021} maximize the information between two heads by enforcing the empirical cross-correlation between the embeddings of both heads to be equivalent to the identity matrix. Additionally, VICReg~\cite{Bardes2022} incorporates regularization terms to prevent information collapse, particularly in scenarios involving multimodal data. Therefore, VANP leverages VICReg to learn visual features by maximizing the information between different modalities.

\textbf{Pre-training for Better Representation:} Codevilla \emph{et al.}~\cite{Codevilla2019} demonstrated the value of pre-trained models for training better policies in autonomous vehicles. Subsequently, many works adopted pre-trained computer vision models, often trained on ImageNet~\cite{Dosovitskiy2017, Codevilla2019, Chen2020c, Ohn-Bar2020, Nazeri2021, Chen2021b, Jaeger2023a}. However, general-purpose "foundation models" pre-trained on pretext tasks can achieve richer representations, enabling them to generalize to various downstream tasks with minimal data in a zero- or few-shot manner~\cite{Bengio2013, Payandeh2023, Balestriero2023}. 

The literature has extensively studied foundation models for robot manipulation~\cite{Dadashi2022, Luo2023, Huang2023c, Brohan2023, DiPalo2023a, Hiranaka2023}. For example, R3M~\cite{Nair2023} pre-trained a general visual encoder for manipulation tasks on the Ego4D human video dataset~\cite{Grauman2022}, while CLIPort~\cite{Shridhar2022} leveraged the CLIP model~\cite{Radford2021} to enable language instructions for manipulation.  Dadashi \emph{et al.} proposed AQuaDem~\cite{Dadashi2022}, a framework to learn quantized actions from demonstrations in continuous action spaces, while VANP is doing the opposite by learning visual features from continuous action spaces. Luo \emph{et al.}~\cite{Luo2023} improved AQuaDem by using VQ-VAE~\cite{VanDenOord2017} for offline reinforcement learning. Huang \emph{et al.}~\cite{Huang2023c} proposed Skill Transformer to learn long-horizon robotic tasks with the help of Transformers~\cite{Vaswani2017}. 

Inspired by Taskonomy~\cite{Zamir2018}, Shen \emph{et al.} proposed conditioning visual demonstrations like segmentation and depth maps on actions during fusion rather than employing a naive fusion approach~\cite{Shen2019}. Yang \emph{et al.}~\cite{Yang2023h} projected the visual cues for navigation on the image space and then trained a policy on the augmented image. STERLING~\cite{Karnan2023} and CAHSOR~\cite{Pokhrel2024} have explored the concept of human preference learning and competence-awareness in the context of off-road navigation using SSL. These methods aligned sensor and visual embeddings by maximizing the mutual information between embeddings by leveraging VICReg~\cite{Bardes2022} and BarlowTwins\cite{Zbontar2021} respectively. 

The work by Eftekhar \emph{et al.}~\cite{Eftekhar2023} presented the closest approach to VANP, employing a learnable codebook module to selectively filter visual observations based on the specific task. However, relying on task-relevant information, e.g., picking up the key, requires additional information that is not available without human annotation or using a simulator while VANP does not need access to such information to learn visual features. Wang \emph{et. al.} used noise to pre-train a visual encoder by predicting the scale of a patch within the noise image that applies to crop the goal observable from the current frame in real experiments~\cite{Wang2023visual}. In contrast, VANP deliberately disregards such task-specific information, focusing instead on extracting general navigation-relevant features. Another work closely related to VANP is NavFormer~\cite{Wang2024}, which utilized BYOL~\cite{Grill2020} on two input images retrieved from a simulator. These images differ in the presence of dynamic objects within the scene. However, this approach confines NavFormer to the simulated environment, limiting its applicability to simulation environments where we have full control of the environment, e.g., making objects invisible to learn the importance of the presence and absence of the object as an obstacle. Conversely, VANP achieves real-world data generalization without relying on the pre-definition of specific rules only possible in simulation or through human annotation.

\textbf{Pre-training for Better Policies:}
Foundation models hold promise for learning not only rich representations but also policies that can generalize across robotic tasks. For instance, SayCan~\cite{Ahn2022} integrates pre-trained language skills with robot actions, demonstrating the potential of pre-training for robotic tasks. This allows robots to physically execute tasks, while the language model provides high-level task insights. Evaluations of real-world robotic tasks confirm the effectiveness of this grounded approach in handling abstract, long-duration instructions for a mobile manipulator.  Li \emph{et al.}~\cite{Li2022e} pre-trained language models to initialize policy networks predicting actions. Reid \emph{et al.}~\cite{Reid2022} fine-tuned pre-trained sequence models on offline reinforcement learning tasks as the policy backbone. VPT~\cite{Baker2022} used pseudo-labeled Minecraft YouTube videos to learn a behavior cloning policy that can craft diamonds. VPT learns the inverse dynamics while VANP uses dynamics to learn visual features. GNM~\cite{Shah2022a} learned a general policy to drive any robot by combining multiple datasets of different robot types. ViNT~\cite{Shah2023} further improved GNM by replacing the policy network with a Transformer~\cite{Vaswani2017}. 

%% file: sections/methodology.tex
Learning visual features for robot navigation using only RGB camera input presents several challenges. Unlike traditional approaches that rely on LiDAR or depth cameras, RGB cameras lack explicit geometric information, making navigation more complex~\cite{fox1997dynamic, quinlan1993elastic, xiao2021toward, ravivisually, xiao2022autonomous, xiao2022appl}. Here, we formally define the visual navigation task and the learning setting for Vision-Action Navigation Pre-training (VANP).

\subsection{Problem Definition}

We define visual navigation as the task of navigating an environment with only RGB camera input, as explored in previous works~\cite{Codevilla2018, Codevilla2019, Nazeri2021}. The visual navigation problem can be formalized as follows. \textbf{Input:} The robot is given a sequence of past and current images from its front-facing camera, $o_t = [I_{t-\tau_P}, I_{t-\tau_P + 1}, \ldots, I_t] \in \mathcal{O}$, where $t$ is the current time step, $\tau_P$ is the number of past frames, and $\mathcal{O}$ is the space of all possible image sequences. The robot is also given its current goal e.g., GPS coordinates, pose, image, or next local coordinate in 2D space, $g \in \mathcal{G}$, which determines the direction it should move in the next time step. \textbf{Output:} The robot must select an action $a_t \in \mathcal{A}$ consisting of continuous linear and angular velocities. $\mathcal{A} = [-1, 1]^2$ is the action space, where $[-1, 1]$ maps to the minimal and maximal linear and angular velocity of the robot. \textbf{Visual Navigation:} The goal is to learn a policy, $\pi_{\theta}: \mathcal{O} \times \mathcal{G} \rightarrow \mathcal{A}$, where $\theta$ represents the policy's parameters, to determine which action to take at each time step to reach its goal efficiently while avoiding collisions with others.

\textbf{End-To-End models:} For end-to-end or holistic models, we define the policy $\pi_{\theta}$ as follows: $a = \pi_{\theta}(o, g) = \sigma_{\zeta}(p_{\phi}(o) \oplus q_{\psi}(g))$, where $\sigma$ is the controller policy parametrized by $\zeta$, $p$ is the image encoder parameterized by $\phi$, $q$ is the goal encoder parameterized by $\psi$, and $\oplus$ is the aggregation of two vectors. To learn these parameters, two common approaches are (1) to learn all of them together in an end-to-end manner which makes the training difficult and time-consuming or (2) to pre-train the image encoder separately and only fine-tune the goal encoder along with the controller to reduce training time.

\textbf{Challenges in visual feature learning:} While extensive research has explored learning visual features for computer vision tasks using SSL~\cite{He2020, Chen2020b, Zbontar2021, Bardes2022, Grill2020, Chen2020a}, adapting these models to specific tasks presents unique challenges~\cite{Karnan2023, Pokhrel2024}. Images in the real world contain implicit cues for navigation but are sometimes full of redundant information. In the context of visual navigation, one such challenge lies in learning visual features from image sequences without unnecessarily capturing such a redundancy, which may result in ambiguity. Additionally, it is not trivial to extract contrastive learning signals from visual navigation actions for contrastive SSL, e.g., an action appropriate for one scenario may or may not be appropriate for another, or different actions may be appropriate for the same scenario.  For instance, in a scenario where a pedestrian stands in front of the robot, two equally valid actions exist: overtaking from either the left or right side. In such cases, simply negating the angular velocity cannot yield a meaningful negative sample and can introduce ambiguity. Furthermore, employing actions from different sequences as negative samples might not provide pertinent information for visual navigation, as actions are inherently influenced by the observed environment. In the next section, we show how VANP addresses these challenges and trains the image encoder $p$ without a downstream objective function.

\subsection{Vision-Action Model}
VANP leverages VICReg~\cite{Bardes2022} to maximize the information between past observations, a future goal, and future actions while maintaining the information collapse between input heads to train the image encoder $p$. Unlike vision SSL models that work on the joint embedding of augmented images~\cite{Chen2020d, He2020}, VANP correlates the action space $\mathcal{A}$ and goal space $\mathcal{G}$ with the pixel latent space $\mathcal{O}$ as shown in Fig.~\ref{fig:arch}. We define VANP pre-training as follows: We sample a batch of $(I^i_{t-\tau_P:t}, a^i_{t:t+\tau_F}, g^i_t)$ from dataset $\mathcal{D}$, where $i$ is the sample number, $I^i_{t-\tau_P:t}$ is a sequence of past visual observations starting from $t-\tau_P$ and ending at $t$, $a^i_{t:t+\tau_F}$ is a sequence of future actions starting from $t$ and ending at $t+\tau_F$, and $g^i_t$ is the current goal at time $t$ instantiated as an image in the future $I^i_{t+\tau_F}$. $\tau_F$ is the number of frames in the future and $\tau_P$ is the number of frames in the past. We then feed $I^i_{t-\tau_P:t}$ to $p_{\phi}$, typically a CNN, and all the embeddings to a transformer encoder~\cite{Vaswani2017}, as well as $a^i_{t:t+\tau_F}$ to $f_{\xi}$ as part of another transformer encoder, to learn image $Z^i$ and action $Z^a$ embeddings, respectively. Each transformer contains an additional context token to capture the continuous information among frames. We feed $g^i_t$ to $p_{\phi}$ to generate goal embedding $Z^g$. Finally, we use VANP's objective function to learn $\phi$ and $\xi$:

\begin{equation} \label{eq:loss1}
\begin{split}
    \mathcal{L}_{\text{VANP}}(Z^i, Z^g, Z^a) &= \lambda \mathcal{L}_{\text{VICReg}}(Z^i, Z^g) \\
    &+ (1 - \lambda) \mathcal{L}_{\text{VICReg}}(Z^i, Z^a),
\end{split}
\end{equation}
where $\lambda$ is the importance of each term,  
and $\mathcal{L}_{\text{VICReg}}$ is the VICReg objective function~\cite{Bardes2022} defined as:

\begin{equation} \label{eq:loss2}
\begin{split}
    \mathcal{L}_{\text{VICReg}}(Z^1, Z^2) &= \mu^1 s(Z^1, Z^2) \\ 
    &+ \mu^2[v(Z^1) + v(Z^2)] \\
    &+ \mu^3[c(Z^1) + c(Z^2)].
\end{split}
\end{equation}
$s$ is the distance between embedding spaces, $v$ and $c$ are the variance and covariance of each embedding respectively. $\mu^1$, $\mu^2$, and $\mu^3$ are hyper-parameters controlling the effectiveness of each term. Leveraging VICReg's objective function offers the advantage of circumventing the need for negative samples, which, as mentioned above, is challenging to define within the action space for navigation tasks. We also compare VICReg's performance against BarlowTwins used by Nazeri \emph{et al.}~\cite{nazeri2024vanp} and observe that BarlowTwins tends to prioritize redundant scene features over those with greater relevance to navigation resulting in degraded performance.

\input{sections/result_table}

\subsection{Implementation Details}
We implement VANP with PyTorch~\cite{Paszke2019} and the training is performed on a single A5000 GPU with 24GB memory\footnote{\faGithub~\url{https://github.com/mhnazeri/VANP}}.

\textbf{Model architecture:} Considering the limited computation resources onboard most mobile robots, we choose ResNet-50~\cite{He2015} without the classification head as a low-latency image encoder for $p_\phi$ and we call it VANP-50. We use two TransformerEncoders with additional context vectors~\cite{Devlin2019, Dosovitskiy2021, Darcet2023} with four layers and four heads as the final image and action encoders to produce the embeddings of $Z^i, Z^a \in \mathbb{R}^{512}$. Both encoders are followed by MLPs with three layers as the projection heads to generate the final $Z'^i, Z'^a \in \mathbb{R}^{1024}$. We apply the same $p_\phi$ to the goal image to generate $Z^g \in \mathbb{R}^{512}$. A critical challenge arises from the inherent differences in modalities between the two networks generating the embeddings, leading to significant variations in their output ranges. To address this discrepancy and ensure effective integration, we initialize all deep networks using the Kaiming Normal initialization~\cite{He2015} with a mean of zero and a variance of one. In the context of the downstream model, an MLP is appended to the Transformer's context vector to predict trajectories at three and five seconds into the future, enabling the evaluation of how the extracted features influence both short-term and long-term interactions.

\textbf{Optimization:} We use the ADAMW optimizer~\cite{Loshchilov2019} and train the model for 200 epochs with a batch size of 2048 and a learning rate of $5e^{-4}$. We observe that large batch sizes add more variation to the update stage and improve learning. To ensure a fair comparison, all models are trained for 50 epochs using the same optimizer and hyperparameters during downstream training. The sole exception is the end-to-end model, which requires 100 epochs to guarantee convergence.

\begin{figure}[!t]
	\centering
        \includegraphics[width=\linewidth]{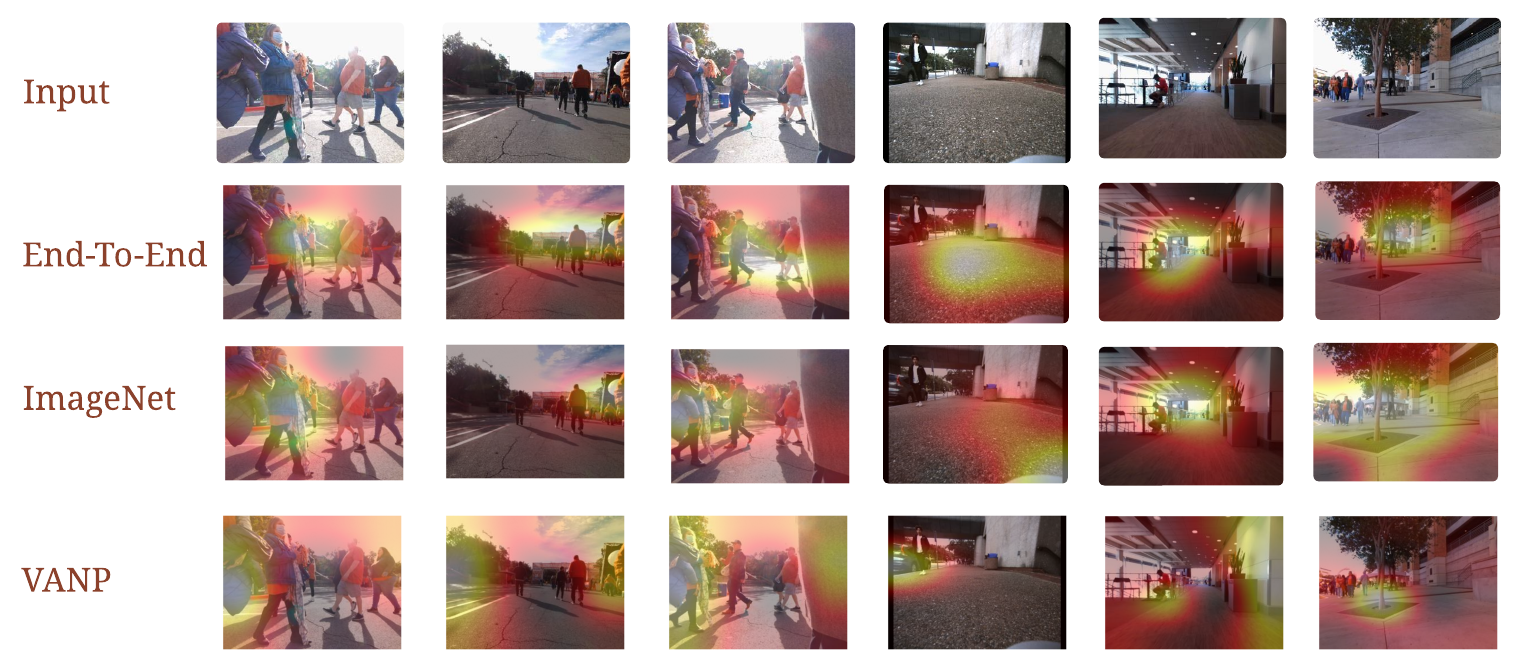}
	\caption{\textbf{Qualitative Comparison.} Comparison of the last layer activation maps among different methods on unseen scenarios.}
	\label{fig:result-qual}
\end{figure}

\textbf{Dataset:} We leverage a selection of two unique datasets: SCAND~\cite{Karnan2022a} and MuSoHu~\cite{Nguyen2023}, both of which encapsulate robot and human navigation data from the egocentric perspective. Both real-world datasets are collected in a variety of natural crowded public spaces. MuSoHu comprises approximately 20 hours of data captured from human egocentric motion. The recordings capture human walking patterns in public spaces, providing insights for learning human-like, socially compliant navigation behaviors. SCAND is an autonomous robot navigation dataset that captures 8.7 hours of human-teleoperated robot navigation demonstrations in naturally crowded public spaces on a university campus. A fundamental limitation of SSL models is their susceptibility to data quality~\cite{Balestriero2023}. As we will discuss in the limitations section (Sec.~\ref{sec:limitations}), VANP is similarly affected, particularly in scenarios where there is no change in a sequence of images as shown in Fig.~\ref{fig:limitations}. To minimize data ambiguity and noise, a subset of the two datasets are carefully curated, ensuring representation of both indoor and outdoor scenes. The resulting dataset, comprising approximately 11,000 samples, was used for both pre-training and training phases. Additionally, a separate set of 8,000 unseen samples are used for downstream navigation task evaluation. For pretext task training, we set $\tau_P$ and $\tau_F$ to 6 and 20 respectively and use a sequence of images $I_{t-\tau_P +1:t} \in \mathbb{R}^{\tau_P \times 98 \times 126}$ along with a goal image $g_t \in \mathbb{R}^{98 \times 126}$ and a sequence of actions $a_{t:t + \tau_F -1} \in \mathbb{R}^{\tau_F \times 2}$ parsed at 4 Hz, comprising of 1.5 seconds in the past and 5 seconds in the future. For the downstream task, we use a sequence of past observations $I_{t - \tau_P +1: t} \in \mathbb{R}^{\tau_P \times 98 \times 126}$ along with the polar coordinates of the next local goal $g \in \mathbb{R}^{2}$ parsed at 4 Hz, containing 1.5 seconds history as the network input to produce the actions $\mathcal{A}_{t:t + \tau_F -1} \in \mathbb{R}^{\tau_F \times 2}$ for three and five seconds in the future.

%% file: sections/result_table.tex
\begin{table*}[t]
\centering
\caption{\textbf{Downstream Performance.} Comparison of the performance of the visual encoders with different pre-training methods on unseen data. Models denoted by an~\faIcon{hourglass-end} require double the training time compared to models with~\faHourglassHalf}~.
\label{tab:results}
\begin{NiceTabular}[columns-width = 0.85cm,rules/width=1pt]{@{}W{l}{25pt}W{l}{5pt}cccccccc@{}}

\toprule                                                
&&       &        &       &       & \Block[c]{1-2}{\textbf{Frozen}~\faSnowflake} & & \Block[c]{1-2}{\textbf{Fine-tuned}~\faFire*} \\
\cmidrule(rl){7-8} \cmidrule(rl){9-10}
\textbf{Type} & & \textbf{Method} & \textbf{Weight} & \textbf{Single-frame} & \textbf{Multiple-frame} & \textbf{3s} & \textbf{5s} & \textbf{3s} & \textbf{5s} \\
\midrule
\Block{2-1} {End-to-End} &\Block{2-1}{~\faIcon{hourglass-end}} &Resnet-50 & Random & \checkmark & \xmark & - & - & 0.116 & 0.307 \\ 
&& ResnetTransformer & Random & \xmark & \checkmark & - & - & 0.113 & 0.320 \\
\midrule
\Block{2-1} {Backbone \\ Supervised} & \Block{2-1}{~\faHourglassHalf}&Resnet-50 & ImageNet & \checkmark & \xmark & \textbf{0.129} & 0.356 & 0.129 & 0.342 \\

&& ResnetTransformer & ImageNet & \xmark & \checkmark & 0.169 & 0.435 & 0.107 & 0.292 \\
\midrule
\Block{2-1} {Backbone \\ Self-Supervised} & \Block{2-1}{~\faHourglassHalf}& Resnet-50 & VANP & \checkmark & \xmark & 0.144 & 0.374 & \textbf{0.103} & \textbf{0.272} \\
&& ResnetTransformer & VANP & \xmark & \checkmark & 0.133 & \textbf{0.342} & 0.114 & 0.319 \\
\bottomrule
\end{NiceTabular}
\end{table*}

%% file: sections/results.tex
We provide experimental results using VANP compared against a ResNet-50 pre-trained on ImageNet and end-to-end from scratch as baselines. 

\subsection{Results Discussion}
We assess the efficacy of VANP pretext training by quantitatively comparing its performance with that of a ResNet-50 model~\cite{He2015} pre-trained on the ImageNet ILSVRC-2012 dataset~\cite{Deng2009}. This serves as the baseline alongside another ResNet-50 model trained end-to-end with randomly initialized weights. To guarantee a fair comparison, the architectures of all other components within the downstream task remain unchanged. Table~\ref{tab:results} presents the mean squared error between the predicted and ground truth trajectories for short- (three seconds) and long-term (five seconds) interactions under two conditions. In the first condition, only the goal encoder and controller are trained during the downstream navigation task, while the image encoder weights are frozen. In the second condition, we compare the performance by unfreezing the image encoder weights to enable fine-tuning.

The results in Table~\ref{tab:results} demonstrate that VANP achieves comparable performance to the end-to-end trained model while requiring only half the training time. Furthermore, VANP pre-trained model achieves comparable performance to ImageNet model with only 0.08\% of the data size required by ImageNet, highlighting how informative the extracted representations are for navigation. 

When provided with a sequence of past observations, VANP exhibits a superior ability (0.342) to utilize this additional data compared to ImageNet model when frozen (0.435). Although the ImageNet weights appear unable to leverage the temporal features provided by the transformer component when freezing its weights (Table~\ref{tab:results}, row four compared against row three), fine-tuning the ImageNet model leads to performance improvement from 0.435 to 0.292, suggesting that it can better capture underlying temporal features provided by the Transformer through fine-tuning. 

However, we do not see such an improvement in the case of VANP. The negligible improvement in accuracy from 0.342 to 0.319 for VANP during fine-tuning can be attributed to two reasons. First, the focus on multiple navigation-related visual regions of VANP's pre-trained weights (Fig.~\ref{fig:result-qual} last row) impedes adaptation/forgetting during fine-tuning compared to the ImageNet weights. Second, the temporal features from the Transformer are already in VANP weights and therefore does not require much fine-tuning. Overall, it is likely that forgetting/updating weights can be easier when the visual encoder is trained using only one single scalar instructive feedback (i.e., training loss) rather than pre-trained on richer instructive signals, i.e., VANP's pre-training objective signal. 

Interestingly, during frozen evaluation with only one image as input, the frozen pre-trained ImageNet model (Table~\ref{tab:results}, row three) achieves the best performance. This finding warrants further investigation. One assumption is that in test cases, the salient object has a stronger influence on the trajectory and aligns better with the single, scalar form of instructive feedback provided. However, during fine-tuning, it is clear that the single image does not outperform models utilizing Transformer temporal features (0.342) while the VANP model benefits from these features even with only one image as input (0.272).

Visual inspection of the learned activation maps on the last layer of ResNet-50 (Fig.~\ref{fig:result-qual}) reveals distinct characteristics across the models. The last row on Fig.~\ref{fig:result-qual} shows that the VANP pre-trained model exhibits activation maps with a higher degree of relevance to navigation tasks, focusing on features such as paths and obstacles while the ImageNet pre-trained model (Fig.~\ref{fig:result-qual} third row) primarily focuses on salient objects within the environment, which might not be directly related to navigation. Another difference between VANP and the end-to-end model (Fig.~\ref{fig:result-qual} second row) is that the end-to-end model tends to concentrate on a single critical region significantly impacting the trajectory, likely due to its limited instructive signal during training, i.e., minimizing the distance between predicted and ground truth trajectories. Conversely, VANP demonstrates the ability to extract information from multiple regions, potentially benefiting from the richer information provided by the goal image and future actions during the pre-training stage. However, as mentioned above, this richness impedes adaptation during fine-tuning. 

We observe instances where the attention of all models shifts to seemingly irrelevant aspects. In the case of VANP, we posit that this may be due to the robot's sharp turns temporarily obscuring the goal image from the current frame.

\begin{figure}
	\centering
        \includegraphics[width=\linewidth]{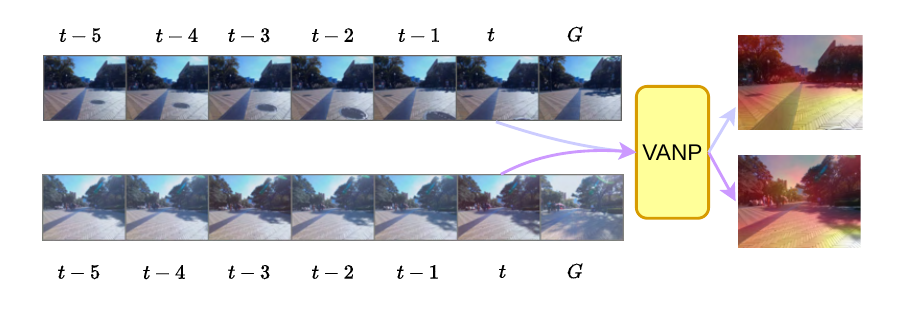}
	\caption{\textbf{Failure Cases.} Samples without any important intra-frame changes cause the model to collapse.}
	\label{fig:limitations}
\end{figure}

\input{sections/ablation_table}

\subsection{Ablations}
To investigate the most effective approach for correlating visual and action spaces, we conduct a series of ablation studies, in which we report the mean squared distance of the predicted trajectory from the ground truth in three and five seconds in the future in Table~\ref{tab:ablation}. 

\textbf{Role of Different Training Signals:} We assessed the individual contributions of various self-supervised training signals by changing the value of $\lambda$ between 0 and 1 in Eq.~\ref{eq:loss1}. Our findings reveal that while action signals provide valuable navigational cues, their sparsity often hinders their effectiveness in downstream navigation tasks, especially during long-term interactions. Conversely, information derived from the goal, while occasionally exhibiting redundancy, improved performance from 0.499 to 0.392 during long-term interactions over using only actions due to informative cues alongside the redundant elements. However, this redundancy poses challenges for the policy network, which can be remedied by more training epochs and a deeper policy network. By combining these two embeddings as the self-supervision signal, the final model can effectively learn informative features while mitigating the impact of redundant information within the embedding.

\textbf{Leveraging Goal Information:} We further investigated the optimal utilization of future goal information. Our findings suggest that employing the goal solely as a supervision signal (shown as Actions+GoalOut in Table~\ref{tab:ablation}) proves more effective in facilitating the model's learning of visual features compared to incorporating the goal directly within the Transformer architecture (shown as Actions+GoalIn in Table~\ref{tab:ablation}). The Transformer's ability to capture temporal changes from the current to the goal frame is only helpful when the goal is visible from the current frame.

\textbf{Augmentations:} Data augmentation is a standard technique employed to enhance model generalization by introducing variability into the dataset. We follow the augmentation scheme outlined by Bardes \emph{et al.}~\cite{Bardes2022} and the result is shown as Augmentations in Table~\ref{tab:ablation}.
We observe that random cropping is particularly critical for VANP, especially in scenarios exemplified by Fig.~\ref{fig:limitations}, as it introduces inter-frame variation. This augmentation strategy relaxes the assumption of carefully curated data and enables an expansion of the dataset from 11,000 to 26,042 samples to include even ambiguous and noisy samples with a little performance hit.

\subsection{Robot Deployment}
To demonstrate the practical applicability of the learned visual features for navigation, a proof-of-concept demonstration of VANP-18 with a moving goal objective~\cite{Xiao2022b} is deployed on a Clearpath Jackal robot. The obstacle avoidance capabilities of VANP are evaluated under controlled conditions. In these experiments, a static obstacle is initially positioned in the robot's path. Subsequent trials involve a dynamic obstacle, simulated by a human pedestrian. Results indicate that VANP exhibits an ability to detect and avoid both static and dynamic obstructions in the majority of test cases. It is important to note that while VANP demonstrates capabilities in object avoidance, it encounters difficulties in navigating around minor obstacles, a limitation likely attributable to restricted visibility conditions. The supplementary video provides a record of these experiments\footnote{\url{https://youtu.be/SEuD9hkwXxQ}}. Despite VANP’s intended versatility across diverse environmental conditions, inherent limitations considering safety only allow it to work in uncluttered environments, as elaborated in the subsequent section.

\subsection{Limitations}\label{sec:limitations} 

We identify multiple key limitations of the VANP pre-training approach. First, our analysis of the learned kernels suggests that VANP performs more effectively when the goal image is directly visible from the current image, likely due to its reliance on image correlation for learning. While this is helpful for Visual-Goal navigation task, it highlights a potential limitation in generalizability to scenarios where the goal location may not be directly visible from the starting point. Second, in large-scale datasets likely with a significant amount of noise, scaling VANP poses a potential challenge, considering its need for high-quality self-supervision during pre-training can result in many changes in learned activation maps between epochs. As can be seen in Fig.~\ref{fig:limitations}, the VANP objective is unable to learn from scenarios where there is no intra-frame change as the time passes. This limitation can be alleviated with augmentations, particularly random cropping, but it does not eliminate it. Additionally, our current findings are based on a static dataset and may not directly translate to challenging real-world navigation tasks that involve dynamic environments and unforeseen obstacles. Further research is needed to evaluate VANP's performance in these more complex scenarios.

%% file: sections/ablation_table.tex
\begin{table}[t]
\centering
\caption{\textbf{Ablations.} Ablation study on the role of each module on the downstream navigation task performance.}
\label{tab:ablation}
\begin{NiceTabular}[columns-width = 1.5cm,rules/width=1pt]{@{}>{\bfseries}lcc@{}}
\toprule
Information & 3s & 5s\\
\midrule
Actions & 0.167 & 0.499 \\
Goal & 0.160 & 0.392 \\
Actions+GoalIn & 0.155 & 0.386 \\
Actions+GoalOut & 0.144 & 0.383 \\
Augmentations & \textbf{0.133} & \textbf{0.342} \\

\bottomrule
\end{NiceTabular}
\end{table}

%% file: sections/conclusion.tex
In this work, we propose a self-supervised learning approach to train visual encoder models specifically designed for visual navigation. This approach is motivated by the observation that humans only pay attention to specific navigation-relevant regions of their frontal view to efficiently make navigation decisions. By reversing this observation, we use the navigation decisions to extract only visual features that are relevant to the navigation task, unlike computer vision models that mainly extract salient details, which are potentially irrelevant to navigation tasks and can therefore lead to confusion for neural-based controllers. To achieve this, we leverage two Transformer Encoders to embed past visual observation, future actions, and a goal image, then we maximize the information between these embeddings using VANP's objective function to learn visual backbone weights.

Furthermore, the VANP objective function facilitates the integration of additional embeddings derived from diverse modalities, including depth data and semantic information or inputs from other sensors such as LiDARs\cite{Panigrahi2023}. Studying the effectiveness of this enrichment of the embedding space with supplementary information for downstream navigation tasks can be a potential future work. Another future direction is to merge datasets from different environments, such as indoor~\cite{liu2021lifelong, atreya2022high}, outdoor~\cite{xiao2021learning, karnan2022vi}, off-road~\cite{datar2023toward, datar2023learning, Datar2024}, and social environments~\cite{xiao2022learning, mirsky2021conflict}, to extend the generalizability of the proposed VANP approach.
More real-world experiments can support all these future directions and scale up the model to larger datasets.